\documentclass[letterpaper, 10 pt, conference]{ieeeconf}  

\usepackage{color}
\newcommand{\ie}{i.e.\ }

\newcommand{\Reffig}[1]{Figure~\ref{#1}}
\newcommand{\Refsec}[1]{Section~\ref{#1}}

\newcommand{\Refeq}[1]{Equation~\ref{#1}}
\newcommand{\Reftab}[1]{Table~\ref{#1}}

\usepackage{graphics} 
\usepackage{epsfig} 
\usepackage{algorithm}
\usepackage{algpseudocode}
\usepackage{amsmath}
\usepackage{amssymb}
\usepackage{subfigure}
\usepackage{graphicx}
\usepackage{subfigure}
\usepackage{booktabs}
\usepackage{threeparttable}
\usepackage{multirow}
\usepackage{multicol}
\usepackage{makecell}
\usepackage{dcolumn}
\newcolumntype{d}[1]{D{.}{.}{#1}}
                                  
\IEEEoverridecommandlockouts
\title{\LARGE \bf
DL-SLOT: Dynamic Lidar SLAM and Object Tracking Based On Graph Optimization
}

\author{Xuebo Tian$^{1,2}$, Junqiao Zhao*$^{1,2,3}$ and Chen Ye$^{1,2}$
\thanks{*This work is supported by the National Key Research and Development Program of China (No. 2018YFB0105103, No. 2018YFB0505400), the National Natural Science Foundation of China (No. U1764261, No. 41801335, No. 41871370)}
\thanks{$^{1}$Department of Computer Science and Technology, School of Electronics and Information Engineering, Tongji University, Shanghai, China}
\thanks{$^{2}$The Key Laboratory of Embedded System and Service Computing, Ministry of Education, Tongji University, Shanghai, China}
\thanks{$^{3}$Institute of Intelligent Vehicles, Tongji University, Shanghai, China}
\thanks{*Corresponding Author:
        {\tt\small zhaojunqiao@tongji.edu.cn}}%
}

\begin{document}

\maketitle
\thispagestyle{empty}
\pagestyle{empty}

\begin{abstract}
Ego-pose estimation and dynamic object tracking are two key issues in an autonomous driving system. 
Two assumptions are often made for them, \ie{the static world assumption of simultaneous localization and mapping (SLAM) and the exact ego-pose assumption of object tracking}, respectively.
However, these assumptions are difficult to hold in highly dynamic road scenarios where SLAM and object tracking become correlated and mutually beneficial.
In this paper, DL-SLOT, a dynamic Lidar SLAM and object tracking method is proposed. 
This method integrates the state estimations of both the ego vehicle and the static and dynamic objects in the environment into a unified optimization framework, to realize SLAM and object tracking (SLOT) simultaneously. 
Firstly, we implement object detection to remove all the points that belong to potential dynamic objects. 
Then, LiDAR odometry is conducted using the filtered point cloud.
At the same time, detected objects are associated with the history object trajectories based on the time-series information in a sliding window.
The states of the static and dynamic objects and ego vehicle in the sliding window are integrated into a unified local optimization framework.
We perform SLAM and object tracking simultaneously in this framework, which significantly improves the robustness and accuracy of SLAM in highly dynamic road scenarios and the accuracy of objects' states estimation. 
Experiments on public datasets have shown that our method achieves better accuracy than A-LOAM.

\end{abstract}

\section{INTRODUCTION}

Lidar Simultaneous Localization and Mapping (SLAM) method has been well studied in recent years as a fundamental capability in autonomous driving vehicles. 
Although many advanced Lidar SLAM methods are proposed and have high accuracy, they all build on the assumption of the static world assumption of SLAM.

In order to eliminate the impact of dynamic objects on the SLAM, \cite{Alireza2016,Azim2012,Dewan2016,moving_landmarks,dynamic_rigid_objects} use prior semantic knowledge to eliminate the point cloud belonging to potential dynamic obstacles directly. 
Nevertheless, the loss of static information in the environment decreases the localization accuracy and even leads to failure. 
\cite{2020dot,bescos2021dynaslam} identify dynamic objects in the environment through object tracking and filter the point cloud of dynamic objects to improve the robustness of SLAM in highly dynamic road scenarios. 
However, object tracking relies on accurate localization results, and dividing the object tracking and SLAM into two independent processes ignores the relationship between SLAM and object tracking. 

This paper proposes DL-SLOT, a dynamic Lidar SLAM and object tracking method, aiming to perform robust and accurate localization and mapping in highly dynamic road scenarios. 
Moreover, the state estimations of the ego vehicle and the static and dynamic objects in the environment are integrated into a unified optimization framework, simultaneously realizing SLAM and object tracking (SLOT).

Firstly, the point cloud captured by Lidar is delivered into an object detector to inference potential dynamic objects, \ie{vehicles, cyclists, and pedestrians}.
At the same time, the Lidar odometry is implemented using the point cloud that filters out the points belonging to the detected objects. 
Then, we conducted data association between the detected objects and the history object trajectories in a sliding window with a fixed time interval, in which time-series information improves the accuracy of association.
Assuming that the object is moving with constant velocity in a short period, we integrate the object state and vehicle pose in the sliding window into a unified local optimization framework. 
By adding constraints into this framework, the states of the static and dynamic objects and ego vehicle can be estimated simultaneously. 

The main contributions of this article include:
\begin{itemize}
\item A unified optimization framework estimating the state of the potential dynamic object and ego vehicle simultaneously.
\item Robust and efficient data association using time-series information in a sliding window with a fixed time interval.
\end{itemize}

\section{RELATED WORKS}

\subsection{Dynamic Lidar SLAM}
The Lidar SLAM method with real-time positioning capability can run stably under the assumption of the static world but fails in highly dynamic scenarios, such as highways and busy urban road because they lack the processing of dynamic obstacles. 
\cite{Victor2019} detects potential dynamic obstacles in the environment and removes the points belonging to obstacles in the point cloud to reduce the impact of dynamic obstacles. 
Based on the filtered point cloud, the odometry can be calculated by frame-to-frame matching. 
Although removing all points of potential dynamic objects can improve the robustness of SLAM, it loses valuable information about static objects in the environment. 
\cite{lonet2019} is a real-time lidar odometry estimation method based on the deep convolutional network. 
In addition to the odometry regression network, \cite{lonet2019} also deploys a dynamic area mask prediction network. 
With predicted dynamic areas, the odometry regression network can pay more attention to static areas to improve the robustness of localization. 
However, due to the lack of supervision in mask prediction network training, it is difficult to accurately estimate the dynamic area in the scene. 
\cite{chen2019iros} uses a neural network to obtain semantic labels of each point. 
Based on the point cloud with semantic marks, this method can construct a global consistent semantic map, and the points belong to potential dynamic objects can be reliably filtered out.
However, the static objects do not contribute to the SLAM optimization.
\cite{DLOAM} first perform ground extraction and non-ground point clustering on point cloud captured by lidar. 
Then, the motion of objects is estimated by the difference between adjacent frames.
The segmented objects are divided into dynamic and static objects based on the motion features, and the stable feature points are extracted from the static objects.
The pose transformation of adjacent frames is solved by matching feature point pairs.

\subsection{SLAM and Object tracking}

SLAM and object tracking are two critical modules in the autonomous driving system. 
Accurate object speed estimation depends on the accurate self-vehicle pose, while SLAM needs to avoid the influence of dynamic objects on localization accuracy by object tracking.

One solution is to divide SLAM and object tracking into two separate processes. 
Firstly, the SLAM process uses static features in the environment to estimate ego-pose. 
The object tracking process is then implemented to estimate the state of the object based on the obtained ego-pose. 
Object tracking based on detection-tracking has been widely used \cite{darms2008vehicle,petrovskaya2009model,cesic2014detection}, and its main steps include data association and status estimation. 
\cite{nnda} is a simple and effective data association method, which selects the detected object closest to the predicted position of the tracking target as the associated object. 
\cite{PDA} solves the problem of single-target data association in a noisy environment, and \cite{mht} deals with the problem of association conflicts. 
\cite{kalman} is the most widely used system state estimation algorithm. 
\cite{extendkalman} deal with more general nonlinear state estimation problems. 

Based on the prior semantic knowledge of the moving object, \cite{Vincent2020object_tracking} uses the deep learning method to perform semantic segmentation in the image. 
After the extended Kalman filter tracks the segmented objects, the dynamic objects are removed from the original depth map, improving the localization and mapping results.

\cite{bescos2021dynaslam} performs pixel-level semantic segmentation on the image and extracts ORB features. 
If a dynamic segmentation instance, such as a car or animal, contains many feature points, the segmentation instance is created as an object, and the key points are marked as dynamic and assigned to the object. 
Under the assumption that the object and camera are moving at a constant speed, the dynamic and static feature points will be matched with the previous frame and the map. 
The objects are associated based on the matched dynamic feature points, and static matched features are used to initialize the camera pose. 
Finally, SLAM and object tracking are realized by optimizing camera pose and object trajectory simultaneously. 
This method considers that object tracking can provide more constraints for SLAM optimization, but the feature-based method limits its ability to track low-texture objects.

\cite{switch_dynamic_slam} is a back-end solution for visual SLAM in  dynamic environment. 
According to the prior information, the number of object observation and the re-projection constraints of object's feature points, potential dynamic objects are devided into ``good" ones and ``bad" ones.
The prior information includes the object motion model, scale information  and prior pose information measured by other approaches.
``good" dynamic objects, static feature points and camera-pose are tightly coupled into a optimization problem, and all the measurement equations are established at one time for the state estimation of the system.
For ``bad" dynamic object, object tracking is implemented based on the optimized camera pose.

\section{METHODS}

\begin{figure}
        \centering
        \includegraphics[width=8cm]{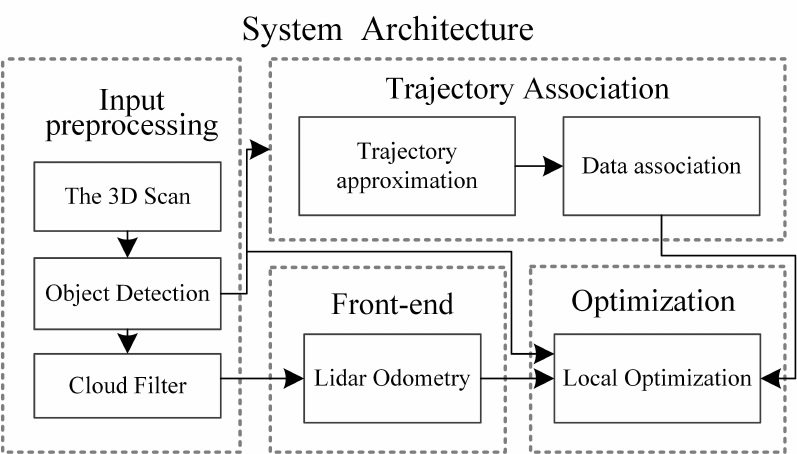}
        \caption{The system architecture of DL-SLOT. The system consists of the input preparation module, the Trajectory association module, optimization modules, and Lidar odometry module. The dashed box indicates the module division, and the rectangle represents the data procession.}
        \label{system}
\end{figure}

As shown in \Reffig{system}, our system mainly includes three modules: trajectory association, optimization and Lidar odometry. 

\subsection{Notations}

We use $T_{t-1}^{t} \subseteq SE(3)$ to represent the transformation of the ego-vehicle from previous frame $t$-1 to the current frame. 
$X_{t} \subseteq SE(3)$ represents the ego-pose at the frame $t$ in the world coordinate system.

The observation result of the $i$-th object from the ego-vehicle at time t is marked as $b_{t}^{i}$. 
$o_{t}^{i}$ represents the pose of this object in the world coordinate system and can be calculated as: 
\begin{equation}
        o_{t}^{i} =  X_{t} \ast b_{t}^{i}
        \label{bti}
\end{equation}

The pose change of the tracked $i$-th object from time $t$-1 to $t$ is $^{i}c_{t-1}^{t}$, which is given by the following formula:

\begin{equation}
        ^{i}c_{t-1}^{t} =  {o_{t-1}^{i}}^{-1} \ast o_{t}^{i}
        \label{ct}
\end{equation}

\subsection{Sliding window based trajectory association}

\begin{figure}
        \centering
        \includegraphics[width=8cm]{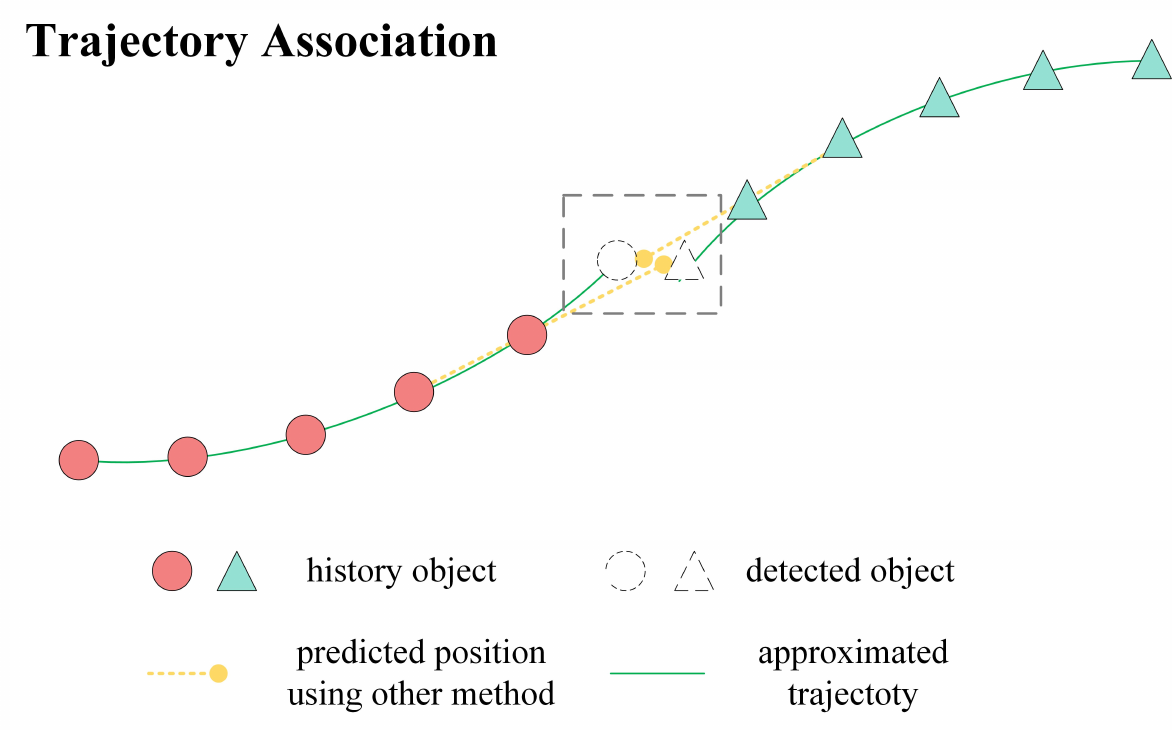}
        \caption{Schematic diagram of trajectory association. The colored circle and triangle represent two different objects. The green solid line represents the approximated object trajectory, and the yellow dashed line represents the Kalman prediction result. The dashed shapes are the detected objects, and the rectangular frame indicates a location that is prone to generate error association. }
        \label{association}
\end{figure}

The object tracking between two frames has some limitations. 
When there are many dynamic objects whose moving trajectories are complicated, such as objects at an intersection with heavy traffic, wrong association results are easily produced. 
For example, as shown in \Reffig{association}, error data association is produced in the area marked as gray dash rectangle, leading to object tracking failure. 
Therefore, a sliding window based trajectory association with better robustness and higher accuracy is proposed.

We maintain a sliding window with a fixed time interval $k$.
Assume that the detected objects set at frame $t$ is $O_{t}=\{ o_{t}^{v} | v=1\cdots V\}$ in the world coordinate system, where $V$ is the number of detected objects.
$O_{t}$ and the historical object trajectories $\{ Tr_{t-1}^{u} | u=1\cdots U \}$ are input into the trajectory association module, where $U$ is the size of object trajectory at frame $t$-1, and $Tr_{t-1}^{u}=\{ o_{t-k}^{u}, \cdots ,o_{t-1}^{u}\}$.

we first approximate the object trajectory $Tr_{t-1}^{u}$ uing object's position of trajectory, as shown in \Refeq{approximate}.

\begin{equation}
        p_{u}(t)=\theta_1 \ast t^{3} + \theta_2 \ast t^{2} + \theta_3 * t + \theta_4
        \label{approximate}
\end{equation}

where $\theta_1$, $\theta_2$, $\theta_3$, $\theta_4$ represent the parameters need to be estimated.
The error between the true position value of trajectory point and the estimated value by approximated function is defined as follows: 

\begin{equation}
        {e_{ap}}_{t}^{u} = {o_{t}^{u}}.pos - p_{u}(t)
        \label{e_approx}
\end{equation}

where $.pos$ represents the position of object $o_{t}^{u}$. 
The parameters are estimated by minimizing the squared sum of the error, as \Refeq{e_approx_all}, using the least-squares problem solution \cite{Li2019}. 

\begin{equation}
        \begin{aligned}
        \underset{\theta}{argmin} \{ &\sum_{i\in [t-k,t-1]} ({e_{ap}}_{i}^{u})^{2} 
                \}
        \end{aligned}
        \label{e_approx_all}
\end{equation}

We approximate the object's trajectory in the x and y axes respectively. 
Using this approximate function, the position of the object at frame $t$ is predicted. 
In order to quantify the degree of matching between trajectory and object, the binary matching score matrix $M_t$ is defined, whose dimension is $(U, V)$.
The element $m_{u, v}=1$ of $M_t$ represents that the $u$-th trajectory and the $v$-th detected object are most likely from the same one and is calculated by follow:
\begin{equation}
        s_{u,v}=\left\{\begin{matrix}
                1&  &dis(p_{u}(t),o_{t}^{v}) < \theta \\ 
                0&  &otherwise
               \end{matrix}\right.
        \label{mvu}
\end{equation}

\begin{equation}
        \begin{aligned}
        dis(p_{u}(t),o_{t}^{v})= &(p_{u}(t).x-o_{t}^{v}.pos.x)^2 + \\
                         &(p_{u}(t).y-{o_{t}^{v}}.pos.y)^2
        \end{aligned}
\end{equation}

Then the Hungarian algorithm is used to solve the data association problem.

%

\subsection{Sliding window based optimization}

Due to the static world assumption of SLAM and the exact ego-pose assumption of object tracking, most autonomous driving systems divide SLAM and object tracking into two independent modules and ignore the relationship between them. 
However, the post of ego-vehicle can be optimized by observing the motion of objects when the state of object is accurately estimated. 
At the same time, the optimized ego-pose help the object tracking module obtaining more accurate state. 
Therefore, we integrate the SLAM and object tracking into a unified optimization framework.

\begin{figure}
        \centering
        \includegraphics[width=8cm]{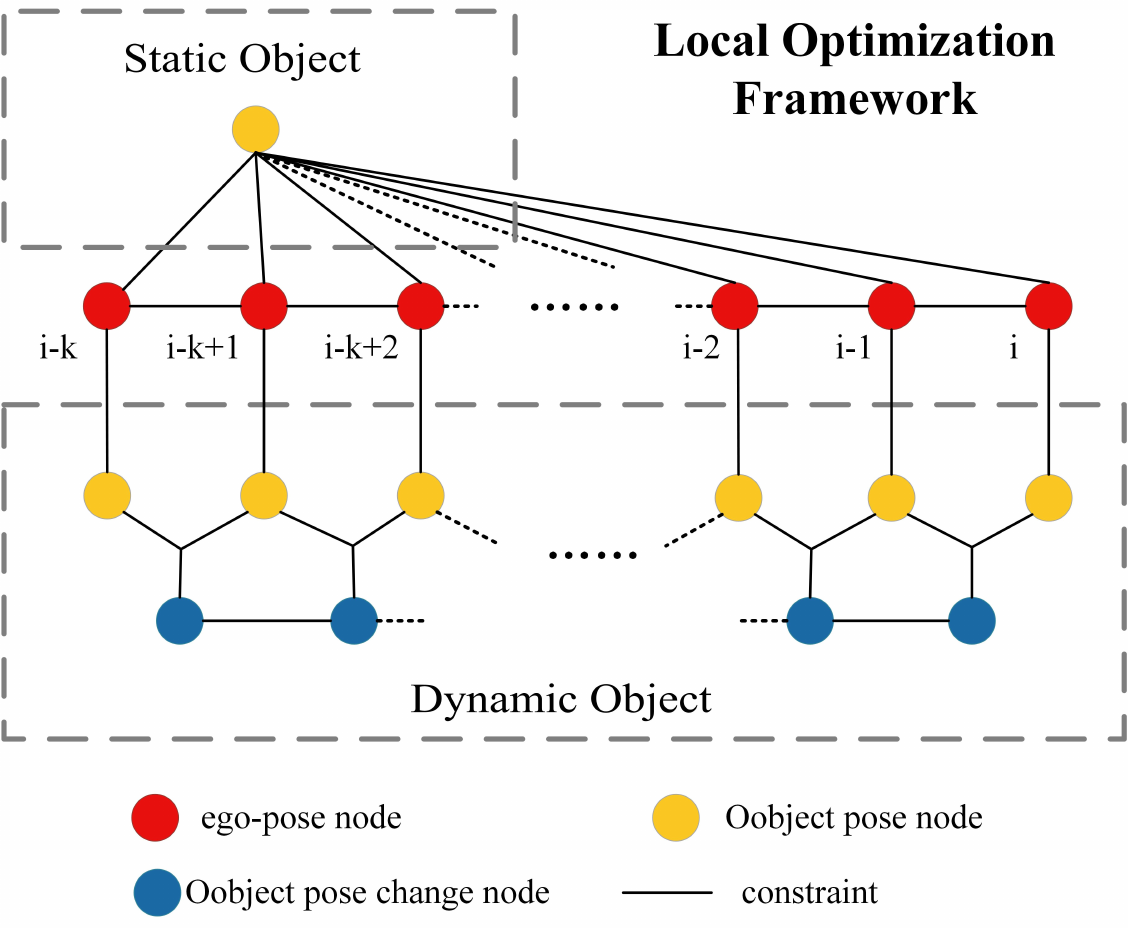}
        \caption{Local optimization framework. The red dot represents the ego-pose vertex, and the yellow dot represents the detected object, and the blue dot represents the pose change vertex of associated object.}
        \label{local_optimization}
\end{figure}

As shown in \Reffig{local_optimization}, all states to be estimated are marked as nodes, and the edges between nodes represent constraints between states. 
Ego-pose $X$ is marked as a red node, and the edge between the two ego-pose nodes represents the Lidar odometry $T$. 
The error between ego-pose nodes and Lidar odometry is defined as follows:
\begin{equation}
        e_{odo}(X_{t-1},X_{k})=({X_{t-1}}^{-1} \cdot {X_{t}})^{-1} \cdot T_{t-1}^{t}
        \label{e_ego}
\end{equation}

When an object $b_{t}$ is detected at $t$ frame, it's global position $o_{t}$ can be calculated by \Refeq{bti}, as shown by the yellow dot in the \Reffig{local_optimization}.
The observation error between detected objects and ego-pose is calculated using \Refeq{e_ego_obj}.
\begin{equation}
        e_{obs}(X_{t},o_{t}^{i},b_{t}^{i})= ({X_{t}}^{-1} \cdot {o}_{t}^{i})^{-1} \cdot {b}_{t}^{i}
        \label{e_ego_obj}
\end{equation}

We add the object pose-change vertex into the optimization graph, marked as yellow dot in \Reffig{local_optimization}.
It's initial value is calculated by \Refeq{ct}, and the error between the object pose and the object pose-change are defined as follows:
\begin{equation}
        e_{chg}(o_{t-1}^{i},o_{t}^{i},^{i}c_{t-1}^{t})= ({o_{t-1}^{i}} \cdot o_{t}^{i})^{-1} \cdot ^{i}c_{t-1}^{t}
        \label{e_obj_obj}
\end{equation}

Support the motion of object is constant speed model in a short period, the error between the pose-change vertex of associated object is given as \Refeq{e_change_change}. 
\begin{equation}
        e_{cons}(c_{t-2}^{t-1},^{i}c_{t-1}^{t})= (^{i}c_{t-2}^{t-1})^{-1} \cdot  ^{i}c_{t-1}^{t}
        \label{e_change_change}
\end{equation}

Finally, \Refeq{e_all} define our optimization problem in the sliding window.
\begin{equation}
        \begin{aligned}
        \underset{}{argmin} \{ &\sum_{i\in [t-k+1,t]} (\left \| e_{odo}(X_{i-1},X_{i}) \right \| ^{2}_{\Sigma_{odo}} + \\
                &\sum_{j\in O^{init}_{i} } \left \| e_{obs}(X_{i},o_{i}^{j},b_{i}^{j}) \right \| ^{2}_{\Sigma_{obs}}+ \\
                &\sum_{j\in O^{aso}_{i}  } \left \| e_{chg}(o_{i-1}^{j},o_{i}^{i},^{i}c_{i-1}^{j}) \right \| ^{2}_{\Sigma_{chg}}+ \\
                &\sum_{j\in O^{cons}_{i} } \left \| e_{cons}(c_{i-2}^{i-1},^{i}c_{i-1}^{i}) \right \| ^{2}_{\Sigma_{cons}} )
                \}
        \end{aligned}
        \label{e_all}
\end{equation}

where $i$ represents the sequence number of the frame in the sliding window, and $\Sigma$ is the covariance matrix. 
$O^{init}$ is the set of objects that have completed the tracking state initialization ,  $O^{aso}$ and $O^{cons}$ respectively represents the set of objects that the number of initialized frames is not less than two and three frames in their trajectory. 
So, their relation is $O^{cons}_{i} \subseteq O^{aso}_{i} \subseteq O^{init}_{i} \subseteq O_{i}$.
When the window is sliding, applying Schur complement trick to the local optimization can reserve the constraint that will be slided out of the window into the current optimization graph. 

%
%

Finally, the graph optimization is implemented based on G2O \cite{grisetti2011g2o}. 

\subsection{System Implementation}
\label{sys_imp}

DL-SLOT is a dynamic Lidar SLAM and Object tracking method. 
The Lidar odometry calculation method of A-LOAM \cite{Li2019} is deployed in our system.
As shown in figure \Reffig{system}, when the system receives a frame of the point cloud, \cite{yan2018second} are firstly implemented to detect the potential dynamic object in the environment. 
Then, the point cloud filtered out the points of detected potential dynamic objects is used for odometry calculation.

A tracking initialization property is set for each object to avoid integrating the false detection into the system. 
If an object is continuously observed more than $g$ frames, its trajectory can be approximated by \Refeq{approximate}, and the approximated trajectory function predicts the object's position at the current time.
The associated detected object with this trajectory is marked as initialized.
For the trajectory having fewer observations, the object's position at the last frame is used to associate with the detected object and mark the associated object as not initialized. 

When the optimization module receives the object detection and association results, the optimization graph is updated. 
Firstly, the pose of the ego-vehicle is added into the optimization graph, and the constraint between adjacent ego-pose is constructed.
Then, the object marked as initialized is integrated into the optimization framework. 
We first use the object's trajectory in the sliding window to preliminarily judge the dynamic or static state of the object. 
For dynamic objects, we add the object pose vertex and object pose change vertex to the local optimization graph, and the constant speed constraint is constructed, as shown in the dynamic object in \Reffig{association}. 
The observation constraint between the existing object pose and the current ego-pose is constructed if it is static. 
Finally, the state sliding out of the window is marginalized, and the optimization is implemented.

\section{EXPERIMENTAL RESULTS}
We conducted the experiment to analyze the effectiveness of the proposed method. 
The public data set KITTI \cite{Geiger2012CVPR} are fully used in our experiment. 

\subsection{Quantitative evaluation}
\label{localization_eva}

\begin{table}[H]
        \caption{RTE and RRE of A-LOAM, A-LOAM* and our method in KITTI dataset.}
        \centering
        \begin{tabular}{cccccccccccc r@{/}}
        \toprule 
        sequence &  A-LOAM & A-LOAM* & ours  \\ 
        \midrule 
        tracking 07&  10.47 / 0.043 & 10.36 / 0.044 & \textbf{ 4.51} / 0.055    \\
        tracking 09&  3.87 / 0.045  & \textbf{3.33} / 0.045  & 3.57 / 0.045    \\
        tracking 13&  \textbf{0.57} / 0.032  & 0.67 / 0.033  & 0.71 / 0.034    \\
        tracking 15&   0.64 / 0.030 &  0.52 / 0.029 & \textbf{ 0.51} / 0.029    \\
        tracking 18&   2.06 / 0.076 &  1.41 / 0.072 & \textbf{ 1.22} / 0.081    \\
        tracking 20&  17.33 / 0.072 &  8.28 / 0.047 & \textbf{ 7.02} / 0.073    \\
        odometry 00&  34.40 / 0.020 & 33.73 / 0.022 & \textbf{24.63} / 0.021   \\
        odometry 05&  20.56 / 0.013 & 20.78 / 0.014 & \textbf{ 8.90} / 0.014    \\
        odometry 08&  35.41 / 0.027 & 34.04 / 0.028 & \textbf{24.06} / 0.027   \\
        odometry 09&  32.76 / 0.020 & 28.56 / 0.020 & \textbf{23.63} / 0.020   \\
        odometry 10&   6.06 / 0.018 &  5.92 / 0.019 & \textbf{ 5.36} / 0.019    \\
        \bottomrule
        \end{tabular}
        \label{rte_and_ree}
\end{table}

\begin{table*}
        \caption{The running time of each module in our system. (the average micro seconds per frame)}
        \centering
        \begin{tabular}{cccccccccc}
          \toprule
            object detection & point filter & A-LOAM odometry & trajectory association & optimization \\
          \midrule
            55 ms&  40s & 84 ms & \textbf{2.3 ms} & \textbf{13.5 ms} \\
          \bottomrule
        \end{tabular}
        \label{time_consume}
\end{table*} 

We selected the KITTI tracking and odometry datasets to prove the effectiveness of our method. 
These datasets are collected in urban areas and highways and contain source point cloud and GPS data. 
We adopted the Relative Translation Error (RTE) metric and the Relative Rotation Error (RRE) metric to evaluate the accuracy of the resulting trajectories. 

Since A-LOAM does not consider the impact of the dynamic objects on localization accuracy, \cite{yan2018second} is implemented to detect potential dynamic obstacles in the point cloud, as mentioned in \Refsec{sys_imp}. 
Furthermore, the point cloud filtering the object's points is delivered into the A-LOAM odometry forming the A-LOAM* result. 
In order to verify the effectiveness of sliding window-based local optimization, we compare the result of A-LOAM , A-LOAM* and DL-SLOT. 
The quantitative results are shown in \Reftab{rte_and_ree}. 

Comparing the experimental results of A-LOAM and A-LOAM* suggests that removing potential dynamic obstacles in the environment improves the performance of the system. 
For example, the tracking sequences contain many highly dynamic scenes, and filtering out the points of potential dynamic objects avoids the selection of unreliable feature points on moving objects. 

The experiment result on our method proves that the sliding window based local optimization significantly improves the localization accuracy. 
In the tracking 20 sequence, the vehicle is driven on the highway and surrounded by moving cars. 
The local optimization improves the positioning accuracy by 1.28 meters, which shows that our optimization strategy makes full use of dynamic obstacles in the environment. 
The scene in the KITTI odometry sequences is primarily static.
After the object tracking state is initialized, the object is regarded as a landmark and added to the local optimization framework, as shown in the static object in \Reffig{association}. 
So, the accuracy of the ego-pose is greatly improved. 
Experiments on the KITTI tracking and odometry sequences show that our sliding window-based local optimization framework is superior in the dynamic and static environment. 

\subsection{Timing analysis}

Our system is deployed on a ubuntu16.04 workstation equipped with an Intel Core i7 3.8 GHz processor, 32G of memory, and GTX2080ti graphics. 
The average computation time of every module in the system is shown in \Reftab{time_consume}. 
The object detection module takes about 55ms to process each point cloud and get the detected bounding box result. 
The time for removing points belonging to the detected object depends on the number of detected objects, and the time of calculation can be further shortened through the parallel operation. 
In this paper, object detection and point filter are run offline, and the saved results are input into the DL-SLOT. 
The sliding window size affects the optimization and tracking time, so we set the size of the sliding window to 10.
Trajectory association and optimization can run with 50fps.
So, they can be embedded in any Lidar SLAM system with a small computational cost.

\section{CONCLUSIONS}

This paper proposed an effection and robust SLAM and object tracking system that is able to operate in dynamic scenes robustly. 
This method integrates the state estimation of surrounding objects and autonomous vehicle into a unified sliding window based optimization framework. 
Therefore, we can perform SLAM and object tracking simultaneously and make these two processes mutually beneficial. 
In addition, we introduce an effective trajectory association method, which can accurately predict the position of an object using the time-series information of the object trajectory.
Our experiments show that DL-SLOT can significantly improve the Localization accuracy in dynamic and static scenarios, which renders our framework applicable to various Lidar SLAM systems. 

In the future, we will explore more efficient and accurate Lidar odometry and loop detection methods and integrate global optimization into our back-end optimization framework to achieve a real-time accurate dynamic Lidar SLAM and object tracking system.

\addtolength{\textheight}{-12cm}
\bibliographystyle{IEEEtran}
\bibliography{reference}

\end{document}